\documentclass{article}
\pdfoutput=1

\usepackage{microtype}
\usepackage{graphicx}
\usepackage{subcaption}
\usepackage{booktabs} 

\usepackage{siunitx}
\usepackage{bm}
\usepackage{bbm}
\usepackage{amsmath}
\usepackage{amssymb}
\usepackage{amsthm}
\usepackage{bm}
\usepackage{comment}
\usepackage{xcolor}
\usepackage{wrapfig}
\usepackage[title]{appendix}

\usepackage{sidecap}

\usepackage{hyperref}

\newcommand{\update}[1]{\color{blue}}

\usepackage[preprint]{icml2021}

\begin{document}

\twocolumn[
\icmltitle{Online hyperparameter optimization by real-time recurrent learning}



\icmlsetsymbol{equal}{*}

\begin{icmlauthorlist}
\icmlauthor{Daniel Jiwoong Im}{nyu}
\icmlauthor{Cristina Savin}{nyu}
\icmlauthor{Kyunghyun Cho}{nyu}
\end{icmlauthorlist}

\icmlaffiliation{nyu}{New York University}

\icmlcorrespondingauthor{Daniel Jiwoong Im}{ji641@nyu.edu}

\icmlkeywords{Machine Learning, ICML}

\vskip 0.3in
]



\printAffiliationsAndNotice{\icmlEqualContribution} 

\begin{abstract}
Conventional hyperparameter optimization methods
are computationally intensive and hard to generalize to scenarios that require dynamically adapting hyperparameters, such as life-long learning.
Here, we propose an online hyperparameter optimization algorithm that is asymptotically exact and computationally tractable, both theoretically and practically. Our framework
takes advantage of the analogy between hyperparameter optimization and parameter learning in recurrent neural networks (RNNs). It adapts a well-studied family of online learning algorithms for RNNs to tune hyperparameters and network parameters simultaneously, without repeatedly rolling out iterative optimization. This procedure yields systematically better generalization performance compared to standard methods, at a fraction of wallclock time.

\end{abstract}

\section{Introduction}

The success of training complex machine learning models critically depends on good choices for hyperparameters that control the learning process. These hyperparameters can specify the speed of learning as well as model complexity, for instance, by setting learning rates, momentum, and weight decay coefficients. Oftentimes, finding the best hyperparameters requires not only extensive resources, but also human supervision.

Well-known procedures, such as random search and Bayesian hyperparameter optimization, require fully training many models to identify the setting that leads to the best validation loss~\citep{Bergstra2012, Snoek2012}. While popular for cases when the number of hyperparameters is small, these methods easily break down when the number of hyperparameters increases. Alternative gradient-based hyperparameter optimization methods take a two level approach (often referred to as the `inner' and the `outer loop'). The outer loop finds potential candidates for the hyperparameters, while the inner loop is used for model training. These methods also require unrolling the full learning dynamics to compute the long-term consequences that perturbation of the hyperparameters has on the parameters~\citep{luketina2016scalable, Lorraine2018, Metz2019}. This inner loop of fully training the parameters of the model for each outer loop hyperparameter update makes existing methods computationally expensive and inherently offline. 
It is unclear how to adapt these ideas to nonstationary environments, as required by advanced ML applications, e.g.\ lifelong learning~\cite{German2019,Kurle2020}. 
Moreover, while these type of approaches can be scaled to high-dimensional hyperparameter spaces~\citep{Lorraine2018}, they can also suffer from stability issues~\citep{Metz2019}, making them nontrivial to apply in practice. 
In short, we are still missing robust and scalable procedures for online hyperparameter optimization. 
 
We propose an alternative class of online hyperparameter optimization (OHO) algorithm that update hyperparameters in parallel to training the parameters of the model. At the core of our framework is the observation that hyperparameter optimization entails a form of \emph{temporal credit assignment}, akin to the computational challenges of training  recurrent neural networks (RNN). Drawing on this similarity allows us to adapt real-time recurrent learning (RTRL)~\citep{Williams1989}, an online alternative to backprop through time (BPTT), for the purpose of online hyperparameter optimization. 

We empirically show that our joint optimization of parameters and hyperparameters yields systematically better generalization performance compared to standard methods. These improvements are accompanied sometimes with a substantial reduction in overall computational cost. We characterize the behavior of our algorithm in response to various meta-parameter choices and to dynamic changes in hyperparameters. OHO can rapidly recover from adversarial perturbations to the hyperparameters. We also find that OHO with layerwise hyperparameters provides a natural trade-off between a flexible hyperparameter configuration and computational cost. In general, our framework opens a door to a new line of research, that is, real-time hyperparameter learning.

Since we have posted the initial version of this preprint, it has been brought to our attention that the early work by~\citep{Franceschi2017} presented a very closely related method to ours, from a different perspective. \citet{Franceschi2017} studied both forward and reverse gradient-based hyperparameter optimization. They derived RTRL \citep{Williams1989} by computing the gradient of a response function from a Lagrangian perspective. In contrast, here we directly map an optimizer to a recurrent network, by taking the process of fully training a model  as propagating a sequence of minibatches through an RNN, which allows us to use RTRL for the purpose of online hyperparameter optimization. Moreover, if \citet{Franceschi2017} included a single small experiment comparing their approach to random search on phonetic recognition, here we empirically evaluate our algorithm's strengths and properties extensively, and compare them to several state of the art methods. 
In doing so, we demonstrate OHO's i) effectiveness: performance w.r.t the wall clock time improves existing approaches, ii) robustness: low variance of generalization performance across parameter and hyparparameter initializations, iii) scalability: sensible trade-off between performance vs. computational cost  w.r.t.\ the number of hyperparameters, and iv) stability of hyperparameter dynamics during training for different choices of meta-hyperparameters.

\section{Background}


\subsection{Iterative optimization}

A standard technique for training large-scale machine learning models, such as deep neural networks (DNN), is stochastic optimization. In general, this takes the form of parameters changing as a function of the the data and the previous parameters, iteratively updating until convergence:
\begin{align}
    \boldsymbol{\theta}_{\tau+1} = g(\boldsymbol{\theta}_{\tau}, \mathbf{B}_{\tau}; \boldsymbol{\varphi}),
    \label{eq:update_rule}
\end{align}
where $\boldsymbol{\theta}$ denotes the parameters of the model, $g$ is the update rule reflecting the learning objectives, ${\boldsymbol \varphi}$ denotes the hyperparameters,  and $\mathbf{B}_{\tau}$ is the random subset of training data (the minibatch) at iteration $\tau$. 

As a representative example, when training a DNN $f$ using stochastic gradient descent, (SGD) learning follows the loss gradient approximated using a random subset of the training data at each time step
\begin{align}
\label{eq:sgd}
    \boldsymbol{\theta}_{\tau+1} \leftarrow \boldsymbol{\theta}_{\tau} - \frac{\alpha}{|\mathbf{B}_\tau|} \sum_{(x,y) \in \mathbf{B}_\tau} \nabla_{\theta} l(y, f(x; \boldsymbol{\theta}_{\tau})),
\end{align}
where $l(\cdot)$ is the per-example loss function and $\alpha$ is the learning rate.
In this case, $g(\cdot)$ is a simple linear update based on the last parameter and the scaled gradient. Other well-known optimizers, such as  RMSprop~\citep{Tieleman2012} or Adam~\citep{Kingma2015}, can also be expressed in this general form.

\subsection{Recurrent neural networks}
\label{sec:rtrl}


An RNN is a time series model. At each time step $t =\{ 1\cdots T\}$, it updates the memory state $\mathbf{h}_t$ and generates the output $\mathbf{o}_{t+1}$ given the input $\mathbf{x}_t$: 
\begin{align}
\label{eq:rnn}
   \left( \mathbf{h}_{t+1}, \mathbf{o}_{t+1} \right) = r(\mathbf{h}_{t}, \mathbf{x}_t; \boldsymbol{\phi}),
\end{align}
where the recurrent function $r$ is parametrized by $\boldsymbol{\phi}$.

Learning this model involves optimizing the total loss over $T$ steps with respect to $\boldsymbol{\phi}$ by gradient descent. BPTT is typically used to calculate the gradient by unrolling the network dynamics and backward differentiating through the chain rule:
\begin{align*}
    \nabla_\phi \mathcal{L} 
                            &= \sum^T_{t=1}  \Big(\sum^T_{s\geq t+1}\frac{\partial \mathcal{L}_{s}}{\partial \mathbf{h}_{t+1}} \frac{\partial \mathbf{h}_{t+1}}{\partial \mathbf{h}_{t}} + \frac{\partial \mathcal{L}_t}{\partial \mathbf{h}_{t}} \Big)
                            \frac{\partial \mathbf{h}_t}{\partial \boldsymbol{\phi}},
\end{align*}
where $\mathcal{L}_t$ is the instantaneous loss at time $t$, and the double summation indexes all losses and all the applications of the parameters.
Because the time complexity for BPTT grows with respect to $T$, it can be challenging to compute the gradient using BPTT when the temporal horizon is lengthy. In practice, this is mitigated by truncating the computational graph (`truncated BPTT').

\subsection{Real-time recurrent learning}

RTRL \citep{Williams1989} is an online alternative to BPTT, which --instead of rolling out the network dynamics-- stores a set of summary statistics of the network dynamics that are themselves updated online as new inputs come in. To keep updates causal, it uses a forward view of the derivatives, instead of backward differentiation, with the gradient computed as
\begin{align}
    \nabla_{\boldsymbol{\phi}} \mathcal{L} &
     = \sum^T_{t=1}  \frac{\partial \mathcal{L}_{t+1}}{\partial \mathbf{h}_{t+1}} \Big(\frac{\partial \mathbf{h}_{t+1}}{\partial \mathbf{h}_{t}}\frac{\partial \mathbf{h}_{t}}{\partial \boldsymbol{\phi}} +\frac{\partial \mathbf{h}_{t+1}}{\partial \boldsymbol{\phi}_t} \Big).
     \label{eq:rtrl}
\end{align}
The Jacobian $\Gamma_{\tau}=\frac{\partial \mathbf{h}_{t}}{\partial \boldsymbol{\phi}}$ (also referred to as the {\em influence matrix}), dynamically updates as
\begin{align*}
    \Gamma_{\tau} = D_\tau \Gamma_{\tau-1} + G_{\tau-1},
\end{align*}
where $D_\tau = \frac{\partial \mathbf{h}_{\tau}}{\partial \mathbf{h}_{\tau-1}}$, and $G_\tau = \frac{\partial \mathbf{h}_{\tau}}{\partial \boldsymbol{\phi}_{\tau-1}}$. 
In this way, the complexity of learning no longer depends on the temporal horizon. However, this comes at the cost of memory requirements $\mathcal{O}(|\boldsymbol{\phi}||\mathbf{h_t}|)$, which makes RTRL rarely used in ML practice.

\section{Learning as a recurrent neural network}

Both RNN parameter learning and hyperparameter optimization need to take into account long-term consequences of (hyper-)parameter changes on a loss. This justifies considering an analogy between the recurrent form of parameter learning (Eq.~\ref{eq:update_rule}) and RNN dynamics (Eq.~\ref{eq:rnn}).

\paragraph{Mapping SGD to a recurrent network.}
To establish the analogy, consider the following mapping:
\begin{align*}
    \text{parameters } \boldsymbol{\theta}_\tau &\rightarrow \text{state } \mathbf{h}_t \\
    \text{batch } \mathbf{B}_\tau &\rightarrow \text{input } \mathbf{x}_t\\
    \text{update rule } g(\cdot) &\rightarrow \text{recurrent function } r(\cdot)\\
    \text{hyper-parameters } \boldsymbol{\varphi} &\rightarrow \text{parameters } \boldsymbol{\phi}.
\end{align*}
Parameters ${\boldsymbol \theta}_\tau$ and recurrent network state $\mathbf{h}_t$ both evolve via recurrent dynamics, $g(\cdot)$ and $r(\cdot)$, respectively. 
These dynamics are each parametrized by hyper-parameters ${\boldsymbol \varphi}$ and parameters ${\boldsymbol \phi}$. 
The inputs $\mathbf{x}_t$ in the RNN correspond to the minibatches of data $\mathbf{B}_\tau$ in the optimizer, and the outputs $\mathbf{o}_t$ to 
$\left\{ 
    f(x; {\boldsymbol \theta}_\tau) 
    \Big|
    x \in B_\tau
    \right\}$, respectively.
In short, we interpret a full model training process as a single forward propagation of a sequence of minibatches in an RNN.


\paragraph{Mapping the validation loss to the RNN training loss.}

The same analogy can be made for the training objectives. We associate the per-step loss of the recurrent network to the hyperparameter optimization objective:
\begin{align*}
    \mathcal{L}(\boldsymbol{\theta}_\tau, \boldsymbol{\varphi}_\tau; \mathcal{D}_\text{val}) = \frac{1}{|\mathcal{D}_{\text{val}}|} \sum_{(\mathbf{x},\mathbf{y}) \in \mathcal{D}_{\text{val}}} l\big(\mathbf{y}, f(\mathbf{x}; \boldsymbol{\theta}_\tau(\boldsymbol{\varphi}_\tau))\big),
\end{align*}
where $l$ is the per-example loss function, with $\boldsymbol{\varphi}_\tau = \lbrace \alpha, \lambda \rbrace$  the set of tune-able hyperparameters, which in our concrete example corresponds to learning rate $\alpha$ and regularizer $\lambda$.
The hyperparameter optimization objective is calculated over the validation dataset $\mathcal{D}_{\text{val}}$.
The total loss over a full training run is $L_{\text{Total}} = \sum_{\tau=1}^{\infty} \mathcal{L}(\boldsymbol{\theta}_\tau, \boldsymbol{\varphi}_\tau; \mathcal{D}_{\text{val}}),$
where there are multiple parameter updates within a run. We use $\infty$ to emphasize that the number of epochs is often not determined {\it a priori}.

Both the hyperparameter optimization objective and the update rule, Eq.~\ref{eq:sgd}, contain the per-example loss $l$. 
The former updates the parameters of RNN using $\mathcal{D}_{\text{val}}$, while the latter updates the state of RNN using training batch ${\bf{B}_{\tau}}$. 
This meta-optimization loss function is commonly used for hyperparameter optimization in practice~\citep{Bengio2000, Pedregosa2016}. Nevertheless, the mapping from the optimizer to RNN and the mapping from the validation loss to RNN training objective is novel and enables us to treat the whole hyperparameter optimization process as an instance of RNN training. 



\section{Online hyperparameter optimization}
\label{sec:oho}

\begin{figure}
    \centering
    \includegraphics[page=5,width=0.75\linewidth]{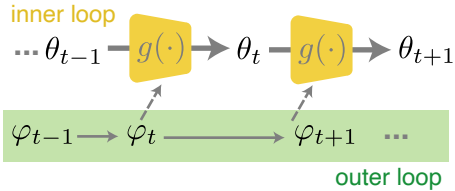}
    \vspace{-0.25cm}
    \caption{Overview of OHO joint  optimization procedure. The parameters and hyperparameters update in parallel based on training/loss gradients computed using the current minibatch.}
    \label{fig:my_label}
    \vspace{-0.5cm}
\end{figure}

With our analogies in place, we are ready to construct our online hyperparameter optimization (OHO) algorithm.
We adapt the hyperparameters at each time step, such that both the parameters and hyperparameters are jointly optimized in a single training process (Fig.~\ref{fig:my_label}).

To achieve this, we update the parameters using training data and update the hyperparameters using validation data.
Then, the update rules for $\boldsymbol{\theta}$ and $\boldsymbol{\varphi}$ are:
\begin{align}
    \boldsymbol{\theta}_{\tau+1} &= \boldsymbol{\theta}_{\tau} - \alpha_{\tau} \Delta_{\mathcal{D}_{\text{tr}}} (\boldsymbol{\theta}_{\tau}) + \alpha_{\tau}  w(\boldsymbol{\theta}_{\tau},\lambda_{\tau}) 
    \label{eq:param_updates}
    \\
    \boldsymbol{\varphi}_{\tau+1} &= \boldsymbol{\varphi}_{\tau} - \eta \Delta_{ \mathcal{D}_{\text{val}}} (\boldsymbol{\varphi}_{\tau}), 
    \label{eq:meta_update}
\end{align}
where $\Delta_{ \mathcal{D}_{\text{tr/val}}}$ is a descent step with respect to training or validation data, respectively, and $w(\boldsymbol{\theta}_{\tau};\lambda_\tau)$ is a regularization function. 
We can use any differentiable stochastic optimizer, such as RMSProp and ADAM, to compute the descent direction $\Delta$. 
Without the loss of generality, from now on, we use SGD to compute the $\Delta$ and use weight decay penalty as a regularizer for the rest of the paper. Expanding $\Delta(\boldsymbol{\varphi}_{\tau}; \mathcal{D}_{\text{val}})$ in Eq.~\ref{eq:meta_update}, we have
\begin{align}
     &\Delta_{\mathcal{D}_{\text{val}}} (\boldsymbol{\varphi}_{\tau}) = \nabla_{\boldsymbol{\varphi}} \mathcal{L}_{\mathcal{D}_{\text{val}}}(\boldsymbol{\theta}_{\tau+1}, \boldsymbol{\varphi}_{\tau}) \nonumber \\
                &= \nabla_{\boldsymbol{\varphi}} \mathcal{L}_{\mathcal{D}_{\text{val}}} \big(\boldsymbol{\theta}_{\tau} -\alpha_{\tau} \big( \Delta_{\mathcal{D}_{\text{tr}}}(\boldsymbol{\theta}_{\tau})- w(\boldsymbol{\theta}_{\tau},\lambda_{\tau})\big), \boldsymbol{\varphi}_{\tau}\big) \label{eq:delta_metaopt}.
\end{align}

The gradient of hyperparameters depends on $\varphi_0\cdots \varphi_\tau$ at iteration $\tau$. 
We apply RTRL to compute this gradient in an online fashion. 
Let us re-write the gradient expression of RTRL in Eq.~\ref{eq:rtrl} taking into account our mappings, 
\begin{align}
     \nabla_{\boldsymbol{\varphi}} \mathcal{L}_{\mathcal{D}_{\text{val}}} &= \sum^T_{\tau=1}  \frac{\partial \mathcal{L}_{\mathcal{D}_{\text{val}}}}{\partial \boldsymbol{\theta}_{\tau+1}} \Bigg( \underbrace{ \frac{\partial \boldsymbol{\theta}_{\tau+1}}{\partial \boldsymbol{\theta}_{\tau}}\frac{\partial \boldsymbol{\theta}_{\tau}}{\partial \boldsymbol{\varphi}} +\frac{\partial \boldsymbol{\theta}_{\tau+1}}{\partial \boldsymbol{\varphi}_{\tau}} }_{\Gamma_{\tau+1}} \Bigg).
     \label{eq:oho_gradient}
\end{align}

Generally, meta-optimization involves computing the gradient through a gradient, as shown in Eq.~\ref{eq:delta_metaopt}. 
This causes the temporal derivative to contain the Hessian matrix,
\begin{align*}
    \frac{\partial \boldsymbol{\theta}_{\tau+1}}{\partial \boldsymbol{\theta}_{\tau}} = I - \alpha_\tau H_\tau - 2 \alpha_\tau\lambda_\tau,
\end{align*}
where $H_\tau=\mathbb{E}_{\mathcal{B}}[\nabla^2_{\boldsymbol{\theta}} \mathcal{L}]$ is the Hessian of the minibatch loss with respect to the parameters.
Then, we plug $\frac{\partial \boldsymbol{\theta}_{\tau+1}}{\partial \boldsymbol{\theta}_{\tau}}$ into the influence matrix's recursive formula,
\begin{align}
    \Gamma_{\tau+1} = \big(I-\alpha_\tau H_{\tau} -2\alpha_\tau\lambda_\tau \big) \Gamma_\tau + G_{\tau}.
    \label{eq:influence_formula}
\end{align}
By approximating the Hessian-vector product 
using the finite difference method,\footnote{
See \url{https://justindomke.wordpress.com/2009/01/17/hessian-vector-products/} for details.
}
we compute the gradient $\frac{d l(\boldsymbol{\theta}_{\tau+1})}{d\alpha_\tau}$ in linear time.
The formulation shows that the influence matrix is composed of all the relevant parts of the learning history and is updated online as new data come in.
In other words, it accumulates the long-term influence of all the previous applications of the hyperparameters $\boldsymbol{\varphi}$ on the parameters $\boldsymbol{\theta}$.

Overall RTRL is more efficient than truncated BPTT for hyperparameter optimization, in both computation and memory complexity.
First, RTRL greatly reduces the computational cost since the influence matrix formulation circumvents the explicit unrolling of temporal dependencies. Our approach is intrinsically online, while BPTT is strictly an offline algorithm (although it can sometimes be loosely approximated to mimic online methods, see~\citep{Lorraine2018}). Moreover, the standard memory challenge of training RNNs with RTRL does not apply in the case of hyperparameter optimization, as the number of hyperparameters is in general much smaller than the number of parameters. For example, one popular choice of $\varphi$ is to just include a global learning rate and some regularization coefficients. 

\subsection{Gradient computation}
\label{sec:gradient_computation}
Since we frame hyperparameter optimization as RNN training, one might reasonably wonder about the fundamental issues of vanishing and exploding gradients~\citep{hochreiter2001gradient} in this setup. Gradients do not vanish, because the SGD update in Eq.~\ref{eq:sgd} is additive. This resembles RNN architectures such as long- short-term memory~\citep[LSTM;][]{hochreiter1997long} and gated recurrent units~\citep[GRU;][]{chung2014empirical}, both of which were explicitly designed to have an additive gradients, as a way of avoiding the vanishing gradient issue. Nonetheless, exploding gradient may still be an issue and is worth further discussion. 

Because the influence matrix $\Gamma_{\tau+1}$ is defined recursively, the gradient of the validation loss with respect to ${\boldsymbol \varphi}$ contains the product of Hessians:
\begin{align*}
{\small 
    \nabla_{\boldsymbol{\varphi}} \mathcal{L}_{\mathcal{D}_{\text{val}}} = \Bigg\langle G_{\tau}, -\sum^{\tau}_{i=0} \Bigg(\prod^{\tau}_{j=i+1} (I-\alpha_j H_j-2\alpha_j\lambda_j)\Bigg) G_i\Bigg\rangle ,
    }
\end{align*}
where $\langle \cdot, \cdot \rangle$ denotes an inner product. $G_i$ and $H_j$ are the gradient and Hessian of the loss $\mathcal{L}(\boldsymbol{\theta}_\tau)$ at iteration $i$ and $j$, respectively. The product of the Hessians can lead to gradient explosion, especially when consecutive gradients are correlated~\citep{Metz2019}.
It remains unclear whether the gradient explosion is an issue for RTRL (or, more generally, forward-differentiation based optimization). We study this in our experiments.
\section{Related work}
Optimizing hyperparameters based on a validation loss has been tackled from various stances. Here, we provide a brief overview of state-of-the-art approaches to hyperparameter optimization.   

\paragraph{Random search.}

For small hyperparameter spaces, grid or manual search methods are often used to tune hyperparameters. 
For a moderate number of hyperparameters, random search can find hyperparameters that are as good or better than those obtained via
 grid search, at a fraction of its computation time \citep{Bengio2000, Bergstra2012, Bergstra2011, Jamieson2015, Li2016}.

\paragraph{Bayesian optimization approaches.}
Bayesian optimization (BO) is a smarter way to search for the next hyperparameter candidate \citep{Snoek2012, Swersky2014, Snoek2015, Eriksson2019, Kandasamy2020} by explicitly keeping track of uncertainty.
BO iteratively updates the posterior distribution of the hyperparameters and assigns a score to each hyperparameter based on it. Because evaluation of hyperparameter candidates must largely be done sequentially, the overall computational benefit from BO's smarter candidate proposal is only moderate.

\paragraph{Gradient-based approaches.}
Hyperparameter optimization with approximate gradient (HOAG) is an alternative technique that iteratively updates the hyperparameters following the gradient of the validation loss.
HOAG approximates the gradient using an implicit equation with respect to the hyperparameters \cite{Domke2012, Pedregosa2016}. 
This work was extended to DNNs with stochastic optimization settings by \citet{Maclaurin2015, Lorraine2020}.  In contrast to our approach, which uses the chain rule to compute the exact gradient online, this approach exploits the implicit function theorem and inverse Hessian approximations. 
While the hyperparameter updates appear online in form, the method requires the network parameters to be nearly at a stable point. In other words, the network needs to be fully trained at each step. 

\citet{Metz2019} attempt to overcome the difficulties of backpropagation through an unrolled optimization process by using a surrogate loss, based on variational and evolutionary strategies. Although their method addresses the exploding gradient problem, it still truncates 
backpropagation \citep{Shaban2019}, 
which introduces a bias in the gradient estimate  of the per step loss. In contrast, our method naturally addresses both issues: It is unbiased by algorithm design, and the gradients are empirically stable, as will be shown in Experiment~\ref{exp:gradient_stability}.

\paragraph{Forward differentiation approaches.}

Our work is closely related to \citep{Franceschi2017, Baydin2018, Donini2020}, which share the goal of optimizing hyperparameters in real-time.
\citet{Franceschi2017} first introduced a forward gradient-based hyperparameter optimization scheme (RTHO), where they compute the gradient of the response function from a Lagrangian perspective. Their solutions ends up being equivalent to computing the hypergradient using RTRL \citep{Williams1989}, as in OHO. Our conceptual framework is however different, as we derive the hyperparameter gradient by directly mapping an optimizer to a recurrent network and then applying RTRL. Unlike \citet{Franceschi2017}, whose only quantitative evaluation was performed on TIMIT phonetic recognition,  we evaluate OHO extensively in a large experimental setup and rigorously analyze its properties. \citet{Donini2020} recently proposed adding a discount factor $\gamma \in [0,1]$ to the influence function, as an heuristic designed to compensate for inner loop non-stationarities by controlling the range of dependencies of the past influence function. 

One potential disadvantage of our approach is that the memory complexity grows with the number of hyperparameters. There are however several approximate variants of RTRL that can be used to address this issue.
As an example, unbiased online recurrent optimization (UORO) reduces the memory complexity of RTRL from quadratic to linear \cite{Tallec2017}.
UORO uses rank-1 approximation of the influence matrix and can be extended to rank-$k$ approximation~\citep{Mujika2018}.
Although UORO gives an unbiased estimate of the gradient, it is known to have high variance, which may slow down or impair training \cite{Cooijmans2019, Marschall2020}. It remains an open question whether these more scalable alternatives to RTRL are also applicable for online hyperparameter optimization.

\begin{figure}[t]
    \centering 
    \includegraphics[width=0.8\linewidth,page=2]{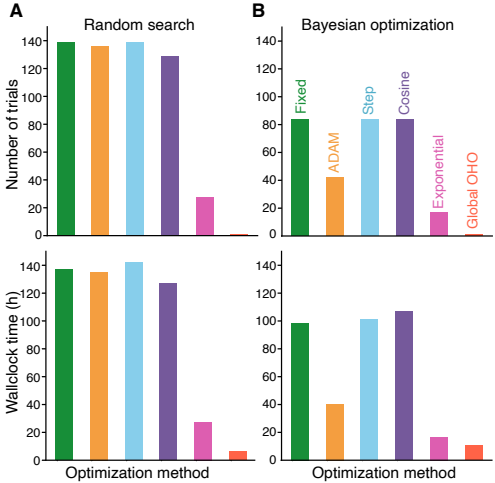}   
    \vspace{-0.25cm}
        \caption{The number of search trials and time required for a model to reach $0.3$ or lower cross-entropy loss on CIFAR10 for various inner-loop learning procedures. Corresponding hyperparameters are selected by 
    (A)  uniform sampling or (B)  Bayesian-optimization (off-the-shelf package 
    scikit-optimize). 
    }
        \label{fig:performance}
        \vspace{-0.25cm}
\end{figure}

\begin{figure}[t]
    \centering 
    \includegraphics[width=0.97\linewidth,page=1]{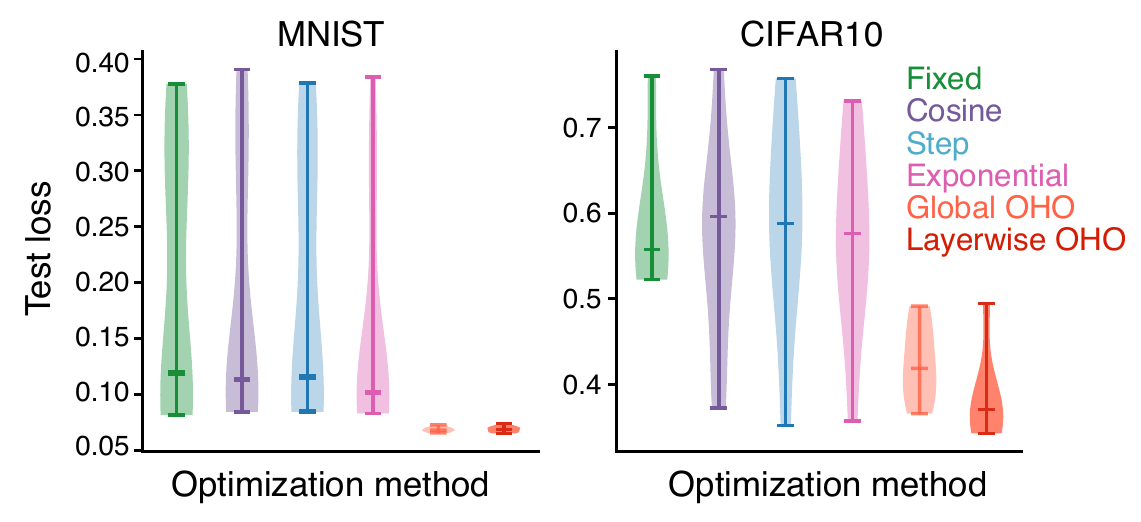}   
    \vspace{-0.25cm}
    \caption{
    Test performance distribution across different hyper-parameters settings (initial learning rates and L2 weight decay coefficients) using different optimization methods.}
    \label{fig:perm_stability}
    \vspace{-0.25cm}
\end{figure}

\section{Experiments}
We conduct empirical studies to assess the proposed algorithm's performance and computational cost. We compare them to widely used hyperparameter optimization methods, on standard learning problems such as MNIST and CIFAR10 classification.  To better understand the properties of OHO, we examine hyperparameter dynamics during training and how they are affected by various meta-hyperparameter settings. We explore layerwise hyperparameter sharing as a potential way to trade-off between computational costs and the flexibility of the learning dynamics. 
Lastly, we analyze the stability of the optimization procedure with respect to meta-hyperparameters. 

For the MNIST~\citep{LeCun1998MNIST} experiments, we use 10,000 out of the 60,000 training examples as the validation set, for evaluating OHO outer-loop gradients. We assess two architectures: a 4-layer neural network and a 4-layer convolutional neural network, with ReLU activations and $5\times5$ kernel size. We use 128 units and 128 kernels at each layer for both networks.
For the CIFAR10 dataset~\citep{Krizhevsky2009CIFAR}, we split the images into 45,000 training, 5,000 validation, and 10,000 test images and normalize
them such that each pixel value ranges between [0, 1].  We use the ResNet18 architecture~\citep{He2016ResNet} and apply random cropping and random horizontal flipping for data augmentation \citep{Shorten2019}. For both MNIST and CIFAR10,  the meta optimizer is SGD, with meta learning rate $0.000005$ and initial weight decay coefficient $0$. We set initial learning rates to $0.001$ and $0.01$, and validation batch size to $100$ and $1000$ for MNIST and CIFAR10, respectively.

\subsection{Performance}
We compare OHO against hyperparameter optimization by uniform random search and Bayesian optimization, in terms of
 the number of training runs and the computation time required to achieve a predefined level of generalization performance (test loss $\leq0.3$, see Fig.~\ref{fig:performance}). 
 We vary the optimizer $g$ (Eq.~\ref{eq:update_rule}) by considering five commonly used learning rate schedulers: 1) SGD with fixed learning rate (`Fixed'), 2) Adam, 3) SGD with step-wise learning rate annealing (`Step'), 4) exponential decay (`Exp'), and 5) cosine (`Cosine') schedulers \citep{Loshchilov2017}. 
 We optimize the learning rate and weight decay coefficient, and all models are trained for 100 and 300 epoch for MNIST and CIFAR10, respectively.
Additionally, we tune step size, decay rate, and momentum coefficients for Step, Exp, and Adam, respectively.
In our experiments, we find that OHO usually takes a single run to find a good solution. 
The single training run takes approximately $12\times$ longer than training the network once without OHO, and yet, the other hyperparameter algorithms require multiple training of neural network to reach the same level of performance.  
Overall OHO ends up significantly faster at finding a good solution in terms of total wall-clock time (Fig.~\ref{fig:performance}, bottom row).

For each method, we measure the final test loss distribution obtained for various initial learning rates and weight decay coefficients.
The initial learning rates and weight decay coefficients were randomly chosen from the range of $[0.0001,0.2]$ and $[0,0.0001]$, respectively. 
For both MNIST and CIFAR10, our procedure, which dynamically tunes the learning rate and degree of regularization during model learning, results in better generalization performance and much smaller variance compared to other optimization methods (Fig.~\ref{fig:perm_stability}, `global OHO'). 
This demonstrates the robustness of OHO to the initialization of hyperparameters and makes it unnecessary to train multiple models in order to find good hyperparameters. 
These results also suggest that there may be a systematic advantage in jointly optimizing parameters and hyperparameters. 


\begin{figure}[t]
    \centering 
    \includegraphics[width=\linewidth,page=3]{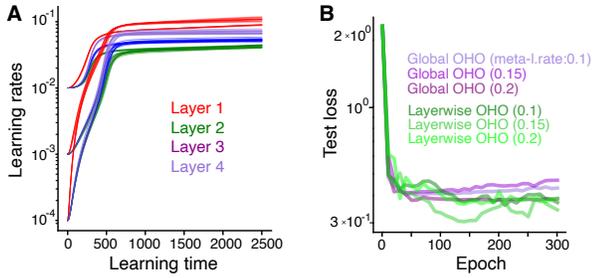}   
    \vspace{-0.5cm}
    \caption{(A) The learning rate dynamics for each layer during the early stage of the training for different initializations.
    (B) Test loss comparison for layerwise vs.\ global OHO, for several initial learning rates.}
    \label{fig:layewise_lr}
    \vspace{-0.25cm}
\end{figure}

\begin{figure}[t]
    \centering 
    \includegraphics[width=0.9\linewidth,page=4]{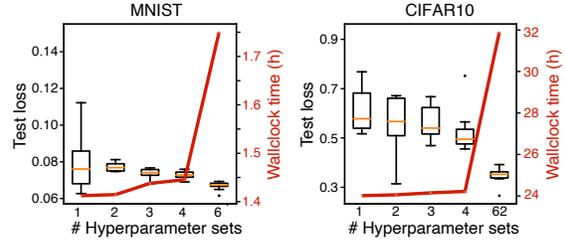} 
    \vspace{-0.25cm}
    \caption{Test loss and wallclock time statistics when using OHO to optimize different numbers of hyper-parameter sets, where each set contains three hyper-parameters:
    two learning rates for weights and bias, and one L2 weight decay coefficient. The hyper-parameter sets are allocated to neighboring layers to evenly partition the full network. Layerwise OHO has 6 sets for MNIST, 62 for CIFAR10. }
    \label{fig:tradeoff_perm_speed}
    \vspace{-0.25cm}
\end{figure}

\begin{figure}[t]
    \centering
    \includegraphics [width=0.9\linewidth,page=7]{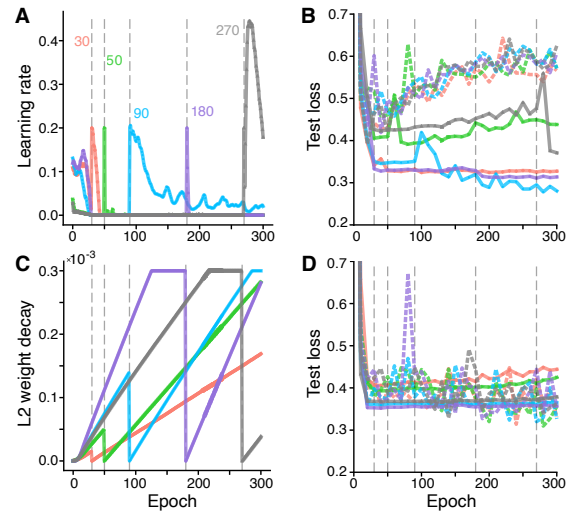}   
    \vspace{-0.25cm}
    \caption{ 
    (A) Learning rate dynamics, after being manually re-initialized to 0.2 at epoch 30, 50, 90, 180, and 270 (vertical dashed lines).
    (B) Test loss of dynamic learning rate (solid line) and non-corrected learning rate (dashed line) after the reset.
    (C) Weight decay dynamics, after being manually re-initialized, details as in (a).
    (D) Corresponding test loss on CIFAR10.}
    \label{fig:perm_corruption}
    \vspace{-0.5cm}
\end{figure}

\subsection{Layerwise OHO}
In the earlier experiment (Fig.~\ref{fig:perm_stability}, `layerwise OHO'), we saw that {\em layer-specific} learning rates and weight decay coefficients can lead to even better test loss on CIFAR10. Here we further study richer hyperparametrizations. 
To do so, we examine the effects of layerwise OHO on final test performance and quantify the trade-offs between performance and computational time, as we vary the number of hyperparameters. 
First, we experiment with having a separate learning rate and weight decay coefficient per layer. We train both global OHO and layerwise OHO on CIFAR10 with several initial learning rates, $\alpha_0 = [0.1,0.15,0.2]$, and no initial regularization, $\lambda_0 = 0$, and compare their test losses (Fig.~\ref{fig:layewise_lr}). We find that the layerwise OHO test loss (green curves) is generally lower than that of global OHO (purple curves) is. Moreover, when analyzing the early stage of training across runs with different initial learning rates, we obverse that learning rates tend to cluster according to their layers regardless of the starting points, and their values are layer-specific (Fig.~\ref{fig:layewise_lr}).  This suggests that layerwise grouping of the hyperparameters may be an appropriate thing to do in such models. 

We analyze the performance and computational speed trade-off of layerwise OHO with coarser hyperparmater grouping schemes. 
If we define one hyperparameter set  as the learning rates for the weight and bias, together with the L2 regularization weights, then global OHO has one set, while layerwise OHO has as many sets as there are layers in the network (6 for MNIST and 62 for CIFAR10). We can then interpolate between these extremes. Specifically, for $k$ hyperparameter sets, we partition the network layers into $k$  neighboring groups, each with their own hyperparameter set (Fig.~\ref{fig:tradeoff_perm_speed}). For example, for $k=2$, the 6-layer network trained on MNIST would be partitioned into top 3 and bottom 3 layers, with 2 separate hyperparameter sets. For each configuration, we ran OHO optimization 10 times with different random initializations. 
As shown in Fig.~\ref{fig:tradeoff_perm_speed}, the average performance improves as the number of hyperparameter sets increases, while the variance shrinks. In contrast, the wallclock time increases with the number of hyperparameter sets. This demonstrates a clear trade-off between the flexibility of the learning dynamics (and subsequently generalization performance) and computational cost, which argues for selecting the richest hyperparameter grouping scheme possible given a certain computational budget. 


\subsection{Response to hyperparameter perturbations}

Due to its online nature, the OHO algorithm lends itself to being used for non-stationary problems, for instance when input statistics or task objectives change over time. Here we emulate such a scenario by perturbing the hyperparameters directly under a static learning scenario, and track the evolution of (meta-)learning in response to these perturbations. 
In particular, we initialize the learning rates to a fixed value (0.1) and then reset them to a higher value (0.2) at various points during CIFAR10 training (Fig.~\ref{fig:perm_corruption}A). 
We observe that the learning rates decrease rapidly early on and that after being reset, they drop almost immediately back to the value before the perturbation. This illustrates  the resilience and rapid response of OHO against learning rate corruption which reflects sudden change in an environment. We further compare the performance of OHO to the setting where the learning rate is fixed after re-initialization. Figure~\ref{fig:perm_corruption}B shows the test losses for fixed (dashed line) and OHO (solid line). We find that the dynamic optimization of hyperparameters leads to systematically better performance.

In the next set of experiments we reset the weight decay coefficient to zero at various epochs: 30, 50, 90, 180, and 270.
The results are similar, with  the weight decay coefficients quickly adapting back to their previous values (Fig.~\ref{fig:perm_corruption}C and D). The test loss fluctuates less for continually optimized OHO, relative to the fixed hyperparameters setting. 
Altogether, these results suggest that joint optimization of parameters and hyperparameters is more robust to fluctuations in a learning environment.

\begin{figure}
        \centering 
    \includegraphics[width=\linewidth,page=6]{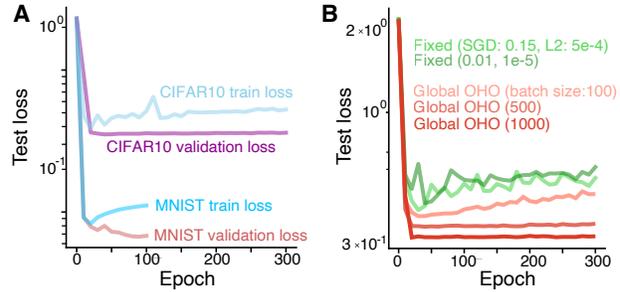}   
    \vspace{-0.5cm}
    \caption{Sensitivity analysis on CIFAR10. A) The comparison of learning when the outer-gradient $\frac{\partial L_{\mathcal{D}}}{\partial \boldsymbol \varphi}$ is subject to training vs validation dataset. B) The performance when using 100, 500, and 1000 validation batch sizes. }
    \label{fig:sensitivity_analysis}
    \vspace{-0.25cm}
\end{figure}


\begin{SCfigure}
\includegraphics[width=0.5\linewidth,page=10]{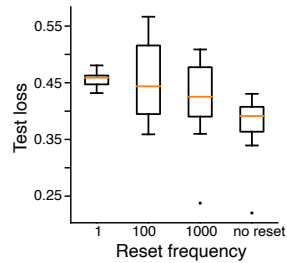}        
\vspace{-0.5cm}
     \caption{Test loss as a function of learning time, when the influence get reset to zero at every 1, 100, and 1000 steps; `no reset' corresponds to standard infinite-horizon global OHO. }
     \label{fig:sensitivity_analysisC}
\end{SCfigure}

\subsection{Long-term dependencies in learning and their effects on hyperparameters } 
Past work has suggested that the temporal horizon for the outer-loop learning influences both performance and the stability of hyperparameter optimization \citep{Metz2019}. In our case, the influence matrix is computed recursively over the entire learning episode and there is no strict notion of a horizon. We can however control the extent of temporal dependencies by resetting the influence matrix at a predefined interval (effectively forgetting all previous experience). We compare such resetting against our regular infinite-horizon procedure (global-OHO; see Fig.~\ref{fig:sensitivity_analysisC}). We find that both the average test loss and its variance decrease as we reset less frequently; ultimately global OHO performs the best. 
Thus, we conclude it is important to take into account past learning dynamics for improving generalization performance.

\subsection{Sensitivity analysis for meta-hyperparameters.}

In order to better understand how OHO works in practice, we explore its sensitivity to meta-hyperparameter choices: the validation dataset, initial learning rates, and meta learning rates. 

\paragraph{Validation dataset.}

We explore the gradient of the meta-optimization objective $\nabla_{\boldsymbol \varphi} \mathcal{L}_{\mathcal{D}_{\text{val}}}$ in Eq.~\ref{eq:oho_gradient}, where the outer gradient $\frac{\partial \mathcal{L}_{\mathcal{D}_{\text{val}}}}{\partial {\boldsymbol \theta}_{\tau+1}}$ is computed with respect to the validation dataset $\mathcal{D}_{\text{val}}$. To show the importance of having a separate validation loss for the outer loop optimization, we replace the validation dataset with the training dataset, and find that it yields to systematically worse generalization performance (Fig.~\ref{fig:sensitivity_analysis}B).  
Expectedly, hyperparameters overfit to the training loss when using the training dataset, but not when using the validation dataset. When using the validation dataset, OHO prevents overfitting by shrinking the learning rates and increasing the weight regularization over time. In other words, OHO early-stops on its own. 

In some practical applications it may prove computationally cumbersome to use the full validation dataset. We experiment using a random subset of validation set to compute the outer gradient while varying the validation minibatch size: 100, 500, and 1,000. Unsurprisingly we find that bigger is better (Fig.~\ref{fig:sensitivity_analysis}C): the test loss is lowest for the largest size considered, while the test loss for the smallest (100) minibatch size is on par with the step-wise learning rate scheduler. The performance gradually changes as we vary the size, which allows us to make trade-off between computational cost and the final performance.

\begin{figure}[t]
    \centering
    \includegraphics[width=\linewidth,page=9]{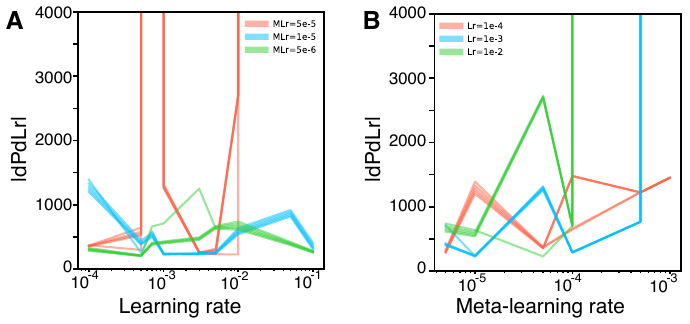}  
    \vspace{-0.5cm}
            \caption{Norm of influence matrix for different meta learning rates and initial learning rates. (A) Large meta learning rates lead to a discontinuous landscape (red curve). (B) The initial learning rates $1e^{-3}$ and $1e^{-2}$ lead to instability when combined with large meta learning rates (green and blue curve).}
    \label{fig:normJ_fix_mlr}
    \vspace{-0.5cm}
\end{figure}

\begin{figure*}[t]
    \centering 
    \includegraphics[width=0.75\linewidth,page=8]{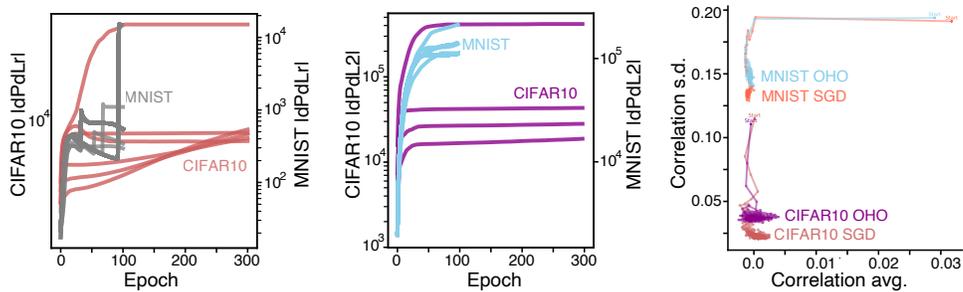}   
    \vspace{-0.5cm}
    \caption{(A,B) Demonstrating that the influence matrix norm w.r.t the learning rate and the weight coefficient plateaus, rather than exploding. (C) The gradients within a single epoch (100 updates) do not correlate with each other. The average and the standard deviation of the correlation decrease over time.}
    \label{fig:stability}
    \vspace{-0.25cm}
\end{figure*}

\paragraph{Meta learning rates and initial hyper-parameters.}

\label{exp:gradient_stability}

To deploy OHO in practice, one needs to choose a meta learning rate and the initial hyperparameters. We investigate the sensitivity of OHO to these choices in order to inform users on what may be a sensible range for each meta-hyperparameter. We define the sensible range to be the region where training is stable. We consider learning to be stable when the gradients are well-defined along the entire learning trajectory, in which computing the gradient hinges upon the product of Hessians  induced by the recursion in the influence matrix. In sum, sensible meta-hyperparameters should keep the norm of the influence matrix bounded throughout the training.  

We visualize the evolution of the norms of the influence matrices for the learning rates, $\|\frac{d{\boldsymbol \theta}^{(T)}}{d\alpha}\|^2_F$, and weight decay coefficients, $\|\frac{d{\boldsymbol \theta}^{(T)}}{d\lambda}\|^2_F$, for different meta learning rates and initial learning rates (Fig.~\ref{fig:normJ_fix_mlr}A, B). The norm is numerical ill-conditioned (blows up) with large meta learning rates ($>1e^{-4}$) at the end stage of training, but is smooth with smaller meta learning rates. 
Recall that the OHO-optimized learning rates decrease during training to avoid overfitting. Later in training this eventually leads to a regime where the meta learning rate is larger than the learning rate. Some elements of the scaled gradient can then become greater than the corresponding learning rate $\alpha_{t-1}^i < \epsilon \left[\nabla_\alpha L({\boldsymbol \theta})\right]^i$, causing instability. The ideal meta learning rate should thus be small.

We further investigate the norm of influence matrix during  training. We train multiple models with different initialization of hyperparameters while fixing the meta learning rate to $5 \times 10^{-5}$.  We find that the norm rarely explodes in practice (Fig.~\ref{fig:stability} A, B). The norms plateau as training converges, but they do not explode. 
We observe similar results for the norm of the Hessian matrix $\|\frac{\partial {\boldsymbol \theta}^{(T)}}{\partial {\boldsymbol \theta}^{(T-1)}}\|^2_2$ as well. 

In order to understand why the norm of influence matrix is stable along the trajectory, we look into exploding gradient phenomena. 
As we discussed in Section~\ref{sec:gradient_computation}, exploding gradients likely happen when a series of gradients are correlated. We compute the moving average and standard deviation of gradient correlations with the window size of 100 updates (a single epoch).
The mean correlation very rapidly approaches zero, with the standard deviation also decreasing as training progresses (Fig.~\ref{fig:stability}C). The gradients only correlate at the beginning of  training but quickly decorrelate as learning continues. This causes the norm of the influence matrix to plateau and prevents the gradients from exploding. 

\section{Discussion}
Truly online hyperparameter optimization remains an open problem in machine learning. In this paper, we presented a novel hyperparameter optimization algorithm, OHO, which takes advantage of online temporal credit assignment in RNNs to jointly optimize parameters and hyperparameters based on minibatches of training and validation data. This procedure leads to robust learning, with better generalization performance than competing offline hyperparameter optimization procedures. It is also competitive in terms of total wallclock time. 

The dynamic interaction between parameter and hyperparameter optimization was found to not only improve test performance but to also reduce variability across runs. Beyond automatic shrinking of learning rates that avoids overfitting, OHO 
quickly adapts the hyperparameters to compensate for sudden changes, such as perturbations of the hyperparameters and allows the learning process to reliably find better models. 

The online nature of OHO updates makes it widely applicable to both stationary and nonstationary learning problems.  We expect the same set of benefits to apply to problems such as life-long learning \citep{German2019,Kurle2020}, or policy learning in deep reinforcement learning \cite{Padakandla2019,Xie2020,Igl2020}, in which the statistics of data and learning objectives change over time. Up to now, it has been virtually impossible to do automated hyperparameter tuning in such challenging learning settings. OHO on the other hand may prove to be a key steppingstone for achieving robust automated hyperparameter optimization in these domains.

In machine learning, an enormous amount of time and energy is spent on hyperparameter tuning. Out of the box, RTRL-based OHO is already quite efficient, with memory requirement that is linear in the number of parameters and outer-loop gradients computation having a similar cost to that of computing inner gradients. This covers many practical use cases of meta-optimization. We believe that OHO and related ideas can dramatically reduce the burden of tuning hyperparameters and becoming a core component of future off-the-shelf deep learning packages.

\nocite{langley00}

\bibliography{main}

\begin{thebibliography}{42}
\providecommand{\natexlab}[1]{#1}
\providecommand{\url}[1]{\texttt{#1}}
\expandafter\ifx\csname urlstyle\endcsname\relax
  \providecommand{\doi}[1]{doi: #1}\else
  \providecommand{\doi}{doi: \begingroup \urlstyle{rm}\Url}\fi

\bibitem[Baydin et~al.(2018)Baydin, Cornish, Rubio, Schmidt, and
  Wood]{Baydin2018}
Baydin, A.~G., Cornish, R., Rubio, D.~M., Schmidt, M., and Wood, F.
\newblock Online learning rate adaptation with hypergradient descent.
\newblock In \emph{International Conference on Learning Representation}, 2018.

\bibitem[Bengio(2000)]{Bengio2000}
Bengio, Y.
\newblock Gradient-based optimization of hyperparameters.
\newblock \emph{Journal of Big Data}, 12:\penalty0 1889--1900, 2000.

\bibitem[Bergstra \& Bengio(2012)Bergstra and Bengio]{Bergstra2012}
Bergstra, J. and Bengio, Y.
\newblock Random search for hyper-parameter optimization.
\newblock \emph{Journal of Machine Learning}, 13:\penalty0 281--305, 2012.

\bibitem[Bergstra et~al.(2011)Bergstra, Bardenet, Bengio, and
  Kegl]{Bergstra2011}
Bergstra, J., Bardenet, R., Bengio, Y., and Kegl, B.
\newblock Algorithms for hyper-parameter optimization.
\newblock In \emph{NeurIPs}, 2011.

\bibitem[Chung et~al.(2014)Chung, Gulcehre, Cho, and
  Bengio]{chung2014empirical}
Chung, J., Gulcehre, C., Cho, K., and Bengio, Y.
\newblock Empirical evaluation of gated recurrent neural networks on sequence
  modeling.
\newblock \emph{arXiv preprint arXiv:1412.3555}, 2014.

\bibitem[Cooijmans \& Martens(2019)Cooijmans and Martens]{Cooijmans2019}
Cooijmans, T. and Martens, J.
\newblock On the variance of unbiased online recurrent optimization.
\newblock In \emph{arXiv preprint arXiv:1902.02405}, 2019.

\bibitem[Domke(2012)]{Domke2012}
Domke, J.
\newblock Generic methods for optimization-based modeling.
\newblock In \emph{International Conference on Artificial Intelligence and
  Statistics}, 2012.

\bibitem[Donini et~al.(2020)Donini, Franceschi, Pontil, Majumder, and
  Frasconi]{Donini2020}
Donini, M., Franceschi, L., Pontil, M., Majumder, O., and Frasconi, P.
\newblock Marthe: Scheduling the learning rate via online hypergradients.
\newblock In \emph{arXiv preprint arXiv:1910.08525}, 2020.

\bibitem[Dougal~Maclaurin(2015)]{Maclaurin2015}
Dougal~Maclaurin, David~Duvenaud, R. P.~A.
\newblock Gradient-based hyperparameter optimization through reversible
  learning.
\newblock In \emph{arXiv preprint arXiv:1502.03492}, 2015.

\bibitem[Eriksson et~al.(2019)Eriksson, Pearce, Gardner, Turner, and
  Poloczek]{Eriksson2019}
Eriksson, D., Pearce, M., Gardner, J.~R., Turner, R., and Poloczek, M.
\newblock Scalable global optimization via local bayesian optimization.
\newblock In \emph{NeurIPs}, 2019.

\bibitem[Franceschi et~al.(2017)Franceschi, Donini, Frasconi, and
  Pontil]{Franceschi2017}
Franceschi, L., Donini, M., Frasconi, P., and Pontil, M.
\newblock Forward and reverse gradient-based hyperparameter optimization.
\newblock In \emph{International Conference on Machine Learning}, 2017.

\bibitem[German et~al.(2019)German, Ronald, Partc, Christopher, and
  Stefan]{German2019}
German, I., Ronald, K., Partc, J.~L., Christopher, K., and Stefan, W.
\newblock Continual lifelong learning with neural networks: A review.
\newblock \emph{Neural Networks}, 113:\penalty0 54--71, 2019.

\bibitem[He et~al.(2016)He, Zhang, Ren, and Sun]{He2016ResNet}
He, K., Zhang, X., Ren, S., and Sun, J.
\newblock Deep residual learning for image recognition.
\newblock In \emph{CVPR}, pp.\  770--778, 2016.

\bibitem[Hochreiter \& Schmidhuber(1997)Hochreiter and
  Schmidhuber]{hochreiter1997long}
Hochreiter, S. and Schmidhuber, J.
\newblock Long short-term memory.
\newblock \emph{Neural computation}, 9\penalty0 (8):\penalty0 1735--1780, 1997.

\bibitem[Hochreiter et~al.(2001)Hochreiter, Bengio, Frasconi, Schmidhuber,
  et~al.]{hochreiter2001gradient}
Hochreiter, S., Bengio, Y., Frasconi, P., Schmidhuber, J., et~al.
\newblock Gradient flow in recurrent nets: the difficulty of learning long-term
  dependencies, 2001.

\bibitem[Igl et~al.(2020)Igl, Farquhar, Luketina, Boehmer, and
  Whiteson]{Igl2020}
Igl, M., Farquhar, G., Luketina, J., Boehmer, W., and Whiteson, S.
\newblock The impact of non-stationarity on generalisation in deep
  reinforcement learning.
\newblock In \emph{arXiv preprint arXiv:2006.05826}, 2020.

\bibitem[Kandasamy et~al.(2020)Kandasamy, Vysyaraju, Neiswanger, Paria,
  Collins, Schneider, Poczos, and Xing]{Kandasamy2020}
Kandasamy, K., Vysyaraju, K.~R., Neiswanger, W., Paria, B., Collins, C.~R.,
  Schneider, J., Poczos, B., and Xing, E.~P.
\newblock Tuning hyperparameters without grad students: Scalable and robust
  bayesian optimisation with dragonfly.
\newblock \emph{Journal of Machine Learning Research}, 2020.

\bibitem[Kevin~Jamieson(2015)]{Jamieson2015}
Kevin~Jamieson, A.~T.
\newblock Non-stochastic best arm identification and hyperparameter
  optimization.
\newblock In \emph{arXiv preprint arXiv:1502.07943}, 2015.

\bibitem[Kingma \& Ba(2015)Kingma and Ba]{Kingma2015}
Kingma, D.~P. and Ba, J.~L.
\newblock Adam: A method for stochastic optimization.
\newblock In \emph{International Conference on Learning Representation}, 2015.

\bibitem[Krizhevsky et~al.(2009)Krizhevsky, Hinton,
  et~al.]{Krizhevsky2009CIFAR}
Krizhevsky, A., Hinton, G., et~al.
\newblock Learning multiple layers of features from tiny images.
\newblock 2009.

\bibitem[Kurle et~al.(2020)Kurle, Cseke, Klushyn, Smagt, and
  Günnemann]{Kurle2020}
Kurle, R., Cseke, B., Klushyn, A., Smagt, P. v.~d., and Günnemann, S.
\newblock Continual learning with bayesian neural networks for non-stationary
  data.
\newblock In \emph{International Conference on Learning Representation}, 2020.

\bibitem[Langley(2000)]{langley00}
Langley, P.
\newblock Crafting papers on machine learning.
\newblock In Langley, P. (ed.), \emph{Proceedings of the 17th International
  Conference on Machine Learning (ICML 2000)}, pp.\  1207--1216, Stanford, CA,
  2000. Morgan Kaufmann.

\bibitem[LeCun et~al.(1998)LeCun, Bottou, Bengio, and Haffner]{LeCun1998MNIST}
LeCun, Y., Bottou, L., Bengio, Y., and Haffner, P.
\newblock Gradient-based learning applied to document recognition.
\newblock \emph{Proceedings of the IEEE}, 86\penalty0 (11):\penalty0
  2278--2324, 1998.

\bibitem[Li et~al.(2016)Li, Jamieson, DeSalvo, Rostamizadeh, and
  Talwalkar]{Li2016}
Li, L., Jamieson, K., DeSalvo, G., Rostamizadeh, A., and Talwalkar, A.
\newblock Hyperband: A novel bandit-based approach to hyperparameter
  optimization.
\newblock In \emph{arXiv preprint arXiv:1603.06560}, 2016.

\bibitem[Lorraine \& Duvenaud(2018)Lorraine and Duvenaud]{Lorraine2018}
Lorraine, J. and Duvenaud, D.
\newblock Stochastic hyperparameter optimization through hypernetworks.
\newblock In \emph{arXiv preprint arXiv:1802.09419}, 2018.

\bibitem[Lorraine et~al.(2020)Lorraine, Vicol, and Duvenaud]{Lorraine2020}
Lorraine, J., Vicol, P., and Duvenaud, D.
\newblock Optimizing millions of hyperparameters by implicit differentiation.
\newblock In \emph{International Conference on Artificial Intelligence and
  Statistics}, 2020.

\bibitem[Loshchilov \& Hutter(2017)Loshchilov and Hutter]{Loshchilov2017}
Loshchilov, I. and Hutter, F.
\newblock Sgdr: stochastic gradient descent with warm restarts.
\newblock In \emph{International Conference on Learning Representation}, 2017.

\bibitem[Luketina et~al.(2016)Luketina, Berglund, Greff, and
  Raiko]{luketina2016scalable}
Luketina, J., Berglund, M., Greff, K., and Raiko, T.
\newblock Scalable gradient-based tuning of continuous regularization
  hyperparameters.
\newblock In \emph{International conference on machine learning}, pp.\
  2952--2960. PMLR, 2016.

\bibitem[Marschall et~al.(2020)Marschall, Cho, and Savin]{Marschall2020}
Marschall, O., Cho, K., and Savin, C.
\newblock A unified framework of online learning algorithms for training
  recurrent neural networks.
\newblock \emph{Journal of Machine Learning}, 21:\penalty0 1--34, 2020.

\bibitem[Metz et~al.(2019)Metz, Maheswaranathan, Nixon, Freeman, and
  Sohl-Dickstein]{Metz2019}
Metz, L., Maheswaranathan, N., Nixon, J., Freeman, C.~D., and Sohl-Dickstein,
  J.
\newblock Understanding and correcting pathologies in the training of learned
  optimizers.
\newblock In \emph{Proceedings of the 36th International Conference on Machine
  Learning (ICML 2020)}, 2019.

\bibitem[Mujika et~al.(2018)Mujika, Meier, and Steger]{Mujika2018}
Mujika, A., Meier, F., and Steger, A.
\newblock Approximating real-time recurrent learning with random kronecker
  factors.
\newblock In \emph{NeurIPs}, 2018.

\bibitem[Padakandla et~al.(2019)Padakandla, J, and Bhatnagar]{Padakandla2019}
Padakandla, S., J, P.~K., and Bhatnagar, S.
\newblock Reinforcement learning in non-stationary environments.
\newblock In \emph{arXiv preprint arXiv:1905.03970}, 2019.

\bibitem[Pedregosa(2016)]{Pedregosa2016}
Pedregosa, F.
\newblock Hyperparameter optimization with approximate gradient.
\newblock In \emph{ICML}, 2016.

\bibitem[Shaban et~al.(2019)Shaban, Cheng, Hatch, and Boots]{Shaban2019}
Shaban, A., Cheng, C.-A., Hatch, N., and Boots, B.
\newblock Truncated back-propagation for bilevel optimization.
\newblock In \emph{International Conference on Artificial Intelligence and
  Statistics}, 2019.

\bibitem[Shorten \& Khoshgoftaar(2019)Shorten and Khoshgoftaar]{Shorten2019}
Shorten, C. and Khoshgoftaar, T.~M.
\newblock A survey on image data augmentation for deep learning.
\newblock \emph{Journal of Big Data}, 6, 2019.

\bibitem[Snoek et~al.(2012)Snoek, Larochelle, and Adams]{Snoek2012}
Snoek, J., Larochelle, H., and Adams, R.~P.
\newblock Practical bayesian optimization of machine learning algorithms.
\newblock In \emph{NeurIPs}, 2012.

\bibitem[Snoek et~al.(2015)Snoek, Rippel, Swersky, Kiros, Satish, Sundaram,
  Prabha, and Adams]{Snoek2015}
Snoek, J., Rippel, O., Swersky, K., Kiros, R., Satish, N., Sundaram, N.,
  Prabha, M. M. A.~P., and Adams, R.~P.
\newblock Scalable bayesian optimization using deep neural networks.
\newblock In \emph{ICML}, 2015.

\bibitem[Swersky et~al.(2014)Swersky, Snoek, and Adams]{Swersky2014}
Swersky, K., Snoek, J., and Adams, R.~P.
\newblock Freeze-thaw bayesian optimization.
\newblock In \emph{arXiv preprint arXiv:1406.3896}, 2014.

\bibitem[Tallec \& Ollivier(2017)Tallec and Ollivier]{Tallec2017}
Tallec, C. and Ollivier, Y.
\newblock Unbiased online recurrent optimization.
\newblock In \emph{arXiv preprint arXiv:1702.05043}, 2017.

\bibitem[Tieleman \& Hinton(2012)Tieleman and Hinton]{Tieleman2012}
Tieleman, T. and Hinton, G.
\newblock Lecture 6.5 - rmsprop, coursera: Neural networks for machine
  learning.
\newblock In \emph{Technical report}, 2012.

\bibitem[Williams \& Zipser(1989)Williams and Zipser]{Williams1989}
Williams, R.~J. and Zipser, D.
\newblock A learning algorithm for continually running fully recurrent neural
  networks.
\newblock \emph{Neural Computation}, 1:\penalty0 270--280, 1989.

\bibitem[Xie et~al.(2020)Xie, Harrison, and Finn]{Xie2020}
Xie, A., Harrison, J., and Finn, C.
\newblock Deep reinforcement learning amidst lifelong non-stationarity.
\newblock In \emph{arXiv preprint arXiv:2006.10701}, 2020.

\end{thebibliography}
\bibliographystyle{icml2021}

\newpage

\end{document}